\documentclass[11pt,a4paper]{article}
\usepackage[hyperref]{emnlp2018}
\usepackage{times}
\usepackage{latexsym}
\usepackage{url}
\usepackage{todonotes}
\usepackage{amsmath}
\usepackage{amsfonts}
\usepackage{graphicx}
\usepackage{subfig}
\usepackage{float}
\usepackage{xcolor, colortbl}
\usepackage{xspace}
\usepackage{booktabs}

\definecolor{blue_uniform}{HTML}{EAF6FF}
\definecolor{blue0}{HTML}{CEEEFF}
\definecolor{blue1}{HTML}{1AB2FF}
\definecolor{blue2}{HTML}{0099E6}

\aclfinalcopy 

\newcommand{\perplexity}[0]{\textsc{Perp}\xspace}
\newcommand{\han}[0]{\textsc{Han}\xspace}

\title{Paying Attention to Attention: Highlighting Influential Samples in Sequential Analysis}

\author{
  Cynthia Freeman \\
  University of New Mexico CS \\
  {\tt cynthiaw2004@gmail.com} \\\And
  Jonathan Merriman \\
  Verint Intelligent Self-Service \\
  {\tt jonathan.merriman@verint.com} \\\AND
  Abhinav Aggarwal \\
  University of New Mexico CS \\
  {\tt abhiag@unm.edu}\\\And
  Ian Beaver \\
  Verint Intelligent Self-Service \\
  {\tt ian.beaver@verint.com} \\\AND
  Abdullah Mueen \\
  University of New Mexico CS \\
  {\tt mueen@cs.unm.edu}}

\date{}

\begin{document}
\maketitle
\begin{abstract}
In~\cite{yang2016hierarchical}, a hierarchical attention network (\han) is created for document classification.  The attention layer can be used to visualize text influential in classifying the document, thereby explaining the model's prediction.  We successfully applied \han to a sequential analysis task in the form of real-time monitoring of turn taking in conversations.  However, we discovered instances where the attention weights were uniform at the stopping point (indicating all turns were equivalently influential to the classifier), preventing meaningful visualization for real-time human review or classifier improvement.  We observed that attention weights for turns fluctuated as the conversations progressed, indicating turns had varying influence based on conversation state.  Leveraging this observation, we develop a method to create more informative real-time visuals (as confirmed by human reviewers) in cases of uniform attention weights using the changes in turn importance as a conversation progresses over time.
\end{abstract}

\section{Introduction}

\begin{table}[h]
\centering
\small{
{
\begin{tabular}{|p{0.6cm}|p{3.5cm}|p{0.9cm}|p{0.9cm}|}
    \hline
    \textbf{Turn} & \textbf{User Text} &  \textbf{\han Weight} & \textbf{Our Weight} \\
    \hline
    \textbf{1} & existing reservation & \cellcolor{blue0}0.33 & 0.0\\ \hline
    \textbf{2} & status & \cellcolor{blue0}0.33 & \cellcolor{blue0}0.33\\ \hline
    \textbf{3} & status of my reservation & \cellcolor{blue0}0.33 & \cellcolor{blue0}0.33\\ \hline
    \textbf{4} & whats the status of reservation \# ggkk98 & \cellcolor{blue0}0.33 & \cellcolor{blue2}1.0\\ \hline
    \textbf{5} & view, assign or change seats & \cellcolor{blue0}0.33 & \cellcolor{blue1}0.66\\ \hline
    \textbf{6} & ggkk98 & \cellcolor{blue0}0.33 & \cellcolor{blue0}0.33\\ \hline
\end{tabular}
}
\caption{The influence on escalation of each user turn in a conversation. Higher weight turns are darker in color. As the \han weights are uniform, and, therefore, similar in color, it is difficult to infer the cause of escalation. In contrast, the weights of our visual on the same conversation show distinct turn importances, thus, quickly indicating the cause of escalation in this conversation.}
\label{tbl:comparison}
}
\end{table}

The attention mechanism~\cite{bahdanau2014neural} in neural networks can be used to interpret and visualize model behavior by selecting the most pertinent pieces of information instead of all available information.  For example, in~\cite{yang2016hierarchical}, a hierarchical attention network (\han) is created and tested on the classification of product and movie reviews. As a side effect of employing the attention mechanism, sentences (and words) that are considered important to the model can be highlighted, and color intensity corresponds to the level of importance (darker color indicates higher importance).

Our application is the escalation of Internet chats. To maintain quality of service, users are transferred to human representatives when their conversations with an intelligent virtual assistant (IVA) fail to progress.  These transfers are known as \textit{escalations}.  We apply \han to such conversations in a sequential manner by feeding each user turn to \han as they occur, to determine if the conversation should escalate. If so, the user will be transferred to a live chat representative to continue the conversation.  To help the human representative quickly determine the cause of the escalation, we generate a visualization of the user's turns using the attention weights to highlight the turns influential in the escalation decision.  This helps the representative quickly scan the conversation history and determine the best course of action based on problematic turns.

Unfortunately, there are instances where the attention weights for every turn at the point of escalation are nearly equal, requiring the representative to carefully read the history to determine the cause of escalation unassisted. Table~\ref{tbl:comparison} shows one such example with uniform attention weights at the point of escalation.


\begin{figure}
\centering \includegraphics[scale = .5]{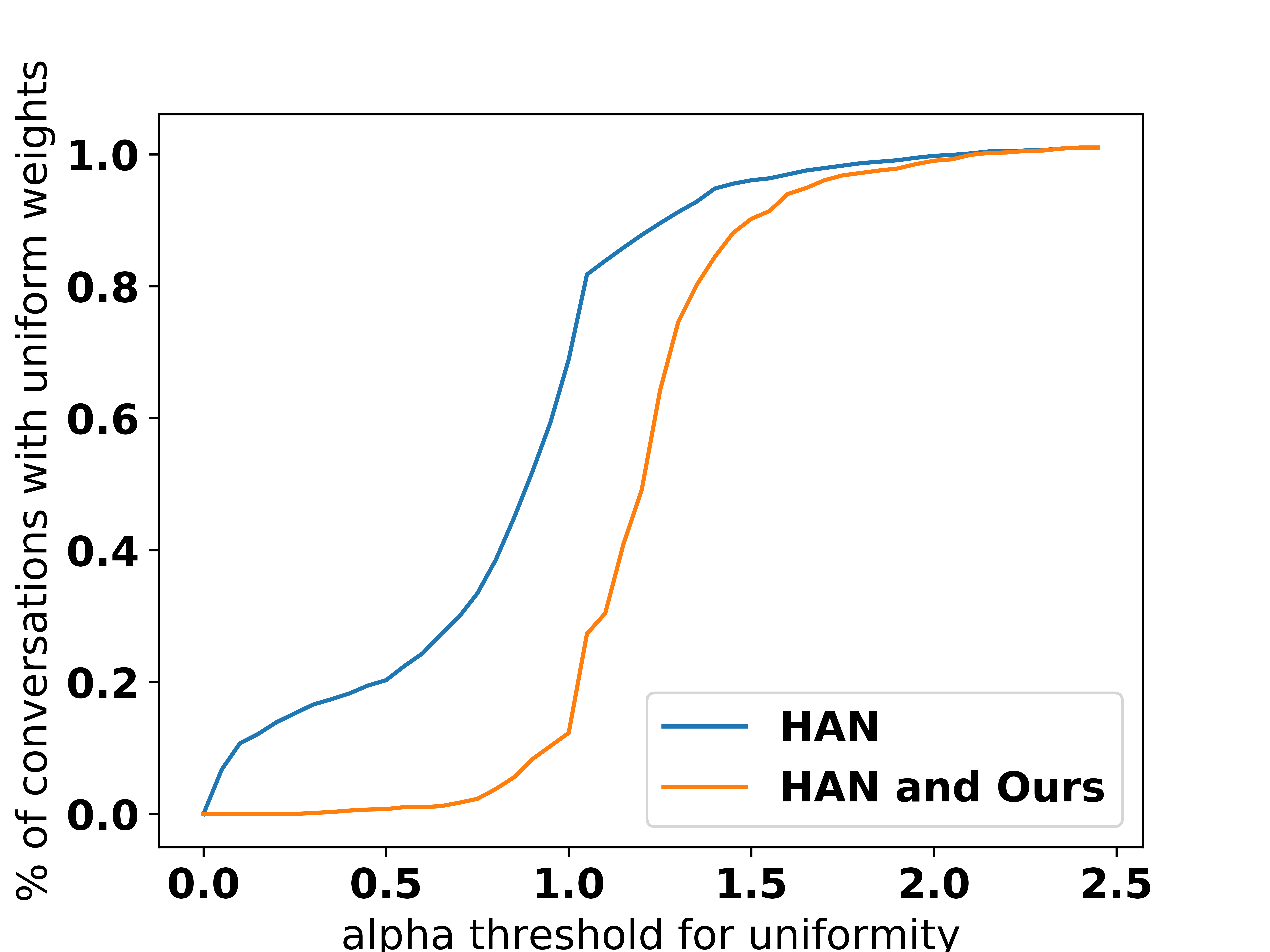}
\caption{The percentage of $1,268$ escalated conversations that would be uniform given a range of thresholds for $\alpha$.  At a threshold of $0.5$, all conversations are considered non-uniform using our method in instances where \han weights are uniform. With \han alone, $20$\% of conversations are uniform at a $0.5$ threshold.}
\label{fig:motivation}
\end{figure}

Our application requires that the visualizations be generated in real-time at the point of escalation. The user must wait for the human representative to review the IVA chat history and resume the failed task. Therefore, we seek visualization methods that do not add significant latency to the escalation transfer. Using the attention weights for turn influence is fast as they were already computed at the time of classification. However, these weights will not generate useful visualizations for the representatives when their values are similar across all turns (see \han Weight in Table~\ref{tbl:comparison}).  To overcome this problem, we develop a visualization method to be applied in the instances where the attention weights are uniform.  Our method produces informative visuals for determining influential samples in a sequence by observing the changes in sample importance over the cumulative sequence (see Our Weight in Table~\ref{tbl:comparison}). Note that we present a technique that only serves to resolve situations when the existing attention weights are ambiguous; we are not developing a new attention mechanism, and, as our method is external, it does not require any changes to the existing model to apply.

To determine when the turn weights are uniform, we use \textit{perplexity}~\cite{brown1992estimate} (see more details in subsection~\ref{subsec:measuring_uniformity}). If a conversation $c$ escalates on turn $i$ with attention weights $[w_1,w_2,...,w_i]$, let $\alpha_c = i - \perplexity(w_1,w_2,...,w_i)$. Intuitively, $\alpha$ should be low when uniformity is high. We measure the $\alpha$ of every escalated conversation and provide a user-chosen uniformity threshold for $\alpha$ (Figure~\ref{fig:motivation}).  For example, if the $\alpha$ threshold for uniformity is $0.5$, $20$\% of conversations in our dataset will result in \han visuals where all turns have similar weight; thus, no meaningful visualization can be produced.  Companies that deploy IVA solutions for customer service report escalated conversation volumes of $1,100$ per day for one IVA~\cite{nitstudies}.  Therefore, even at $20$\%, contact centers handling multiple companies may see hundreds or thousands of conversations per day with no visualizations.  If we apply our method in instances where \han weights are uniform, all conversations become non-uniform using the same $0.5$ threshold for $\alpha$, enabling visualization to reduce human effort.

\section{Related Work}

Neural networks are powerful learning algorithms, but are also some of the most complex. This is made worse by the non-deterministic nature of neural network training; a small change in a learning parameter can drastically affect the network's learning ability. This has led to the development of methodologies for understanding and uncovering not just neural networks, but black box models in general. The interpretation of deep networks is a young field of research.  We refer readers to~\cite{montavon2017methods} for a comprehensive overview of different methods for understanding and visualizing deep neural networks. More recent developments include DeepLIFT~\cite{shrikumar2016not} (not yet applicable to RNNs), layerwise relevance propagation~\cite{bach2015pixel} (only very recently adapted to textual input and LSTMs~\cite{arras2017relevant,arras2017explaining}), and LIME~\cite{ribeiro2016should}. 

LIME is model-agnostic, relying solely on the input data and classifier prediction probabilities. By perturbing the input and seeing how predictions change, one can approximate the complex model using a simpler, interpretable linear model.  However, users must consider how the perturbations are created, which simple model to train, and what features to use in the simpler model. In addition, LIME is an external method not built into the classifier that can add significant latency when creating visuals in real-time as it requires generating perturbations and fitting a regression for every sample point. Attention~\cite{bahdanau2014neural}, however, is built into \han and commonly implemented in other network structures (see below), and, as a result, visuals are created for free as they are obtained from the attention weights directly.

Attention has been used for grammatical error correction~\cite{chopra2016abstractive}, cloze-style reading tasks~\cite{cui2016attention,mihaylov2018knowledgeable}, text classification~\cite{wang2018densely}, abstractive sentence summarization~\cite{ji2017nested}, and many other sequence transduction tasks. ~\cite{yao2015attention} uses an encoder-decoder framework with attention to model conversations and generate natural responses to user input.~\cite{shang2015neural} is perhaps most similar to what we wish to achieve, but only uses one-round conversation data (one user input, one computer response).

To the best of our knowledge, ours is the first paper that considers the changes in attention during sequential analysis to create more explanatory visuals in situations where attention weights on an entire sequence are uniform.

\section{Methodology}

\begin{table}[h]
\centering
\small{
{
\begin{tabular}{|p{1cm}|p{2.5cm}|p{1.2cm}|}
    \hline
    \textbf{Turn} & \textbf{User Text} &  \textbf{\han Weight}  \\ \hline
    \textbf{1} & Hi. I need help. & \cellcolor{blue_uniform} \\ \hline
    \textbf{2} & Is somebody there? & \cellcolor{blue2}\\ \hline
\end{tabular}

\textbf{$\bigg \downarrow$Add turn $3$}

\begin{tabular}{|p{1cm}|p{2.5cm}|p{1.2cm}|}
    \hline
    \textbf{Turn} & \textbf{User Text} &  \textbf{\han Weight} \\ \hline
    \textbf{1} & Hi. I need help. & \cellcolor{blue_uniform} \\ \hline
    \textbf{2} & Is somebody there? & \cellcolor{blue_uniform} \\ \hline
    \textbf{3} & Please help me & \cellcolor{blue_uniform} \\ \hline
\end{tabular}
}
\caption{At the point of turn $2$, \han produces distinct weights.  But at the point of escalation, on turn $3$, the weights are uniform.}
\label{tbl:non-uniform_to_uniform_3_turn_example}
}
\end{table}
In Table~\ref{tbl:non-uniform_to_uniform_3_turn_example}, we see the bottom visualization where the weights are uniform at the point of escalation.  However, on the 2nd turn, the \han had produced more distinct weights.  It is clear from this example that the importance of a single sample can change drastically as a sequence progresses.  Using these changes in attention over the sequence, we formalized a set of rules to create an alternative visualization for the entire sequence to be applied in cases where the attention weights are uniform over all samples at the stopping point.

\subsection{Measuring Uniformity}

\label{subsec:measuring_uniformity}

We begin with defining what it means for attention weights to be uniform. 

For a probability distribution $D$ over the sample space $\Omega$, the perplexity measure is defined as the exponential of the entropy of $D$. More formally, 

$$\perplexity(D) = 2^{H(D)}$$

where the entropy is 

$$H(D) = \sum_{x\in \Omega}D(x)\log_2 \frac{1}{D(x)}$$ 

As  entropy is a measure of the degree of randomness in $D$, perplexity is a measure of the number of choices that comprise this randomness. The following properties of perplexity will be applicable.

\begin{enumerate}
\item \label{perp:positive} For any distribution $D$, the value of $\perplexity(D)$ is always positive. ($2^x > 0$ for all $x$.)
\item \label{perp:upperBound} For any distribution $D$ over $N$ values, we have $\perplexity(D) \le N$. The larger the value, the closer $D$ is to being uniform. The equality holds if and only if $D$ is uniform.
\end{enumerate}
With respect to property (\ref{perp:upperBound}) above, we define a metric $\alpha_N (D) = N - \perplexity(D)$, where $D$ is any distribution over $N$ values. Thus, for all $N \ge 1$ and all distributions $D$ that are uniform over $N$ values, it must be the case that $\alpha_N(D) = 0$. Furthermore, $\alpha_N(D) \ge 0$ for all $N$ and $D$. We drop the subscript $N$ from $\alpha_N(D)$ when it is obvious from the context.

In our application, obtaining an exact uniform distribution is not feasible; it suffices to consider a distribution to be uniform if it is \emph{almost} the same over all values. 

We say that \emph{a given distribution $D$ on $N$ values is $\tau$-uniform if $\alpha_N(D) \leq \tau$}. Note that since $\alpha_N(D)$ can be at most $N-1$ (as $N \ge 1$), this restricts $\tau$ to be any real number between $0$ and $N-1$. 

In this context, given a distribution $D$ over $N$ values, we will refer to $\alpha(D)$ as the \emph{measure of uniformity} of $D$. The smaller the value of $\alpha(D)$, the closer $D$ is to being uniform. 

For our specific application, $\tau$ is a user chosen uniformity threshold, $D$ consists of turn weights, and $N$ is the number of turns in the conversation. For example, in Figure~\ref{fig:motivation}, if the threshold for $\alpha$ is chosen to be $0.5$, this will result in $20$\% of conversations in our datasets with uniform \han turn weights. 

\subsection{Attention Behaviors}

\label{subsec:attn_behaviors}

\begin{figure}
\centering \includegraphics[scale = .33]{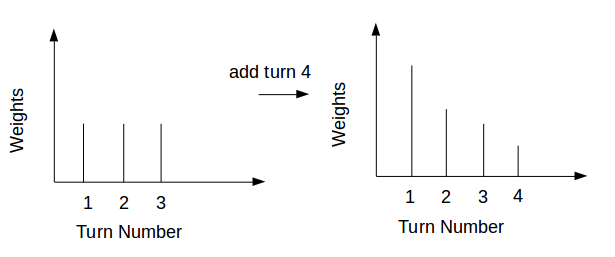}
\caption{An example of an attention dependency switch: Adding turn $4$ caused the distribution of weights to switch from uniform to non-uniform.}
\label{fig:attn_dependency_switch}
\end{figure}

\begin{figure}
\centering \includegraphics[scale = .3]{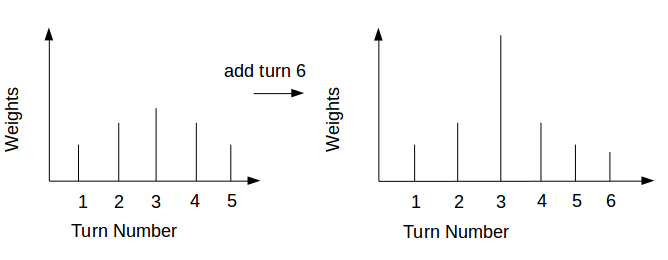}
\caption{An example of a context dependency switch: Adding turn $6$ caused turn $3$'s weight to spike.}
\label{fig:context_dependency_switch}
\end{figure}

Given a conversation $C$ that contains $N$ turns, let $\mathbf{w}_i$ be the vector of attention weights obtained from inputting $T_1,\dots,T_i$ (where $T_i$ is the $i$-th turn in $C$) to \han. When turn $i+1$ is added, we consider three forms of behavior that help us create a new visual: attention, context, and variation dependency switches. See section~\ref{sec:results_and_discussion} for evidence as to why we chose these particular behaviors.

An \emph{attention dependency switch} occurs when the addition of a turn changes the distribution of weights. Suppose we have a $4$ turn conversation. In Figure~\ref{fig:attn_dependency_switch}, considering only the first $3$ turns gives us a uniform distribution of weights (left). However, when we add turn $4$ (Figure~\ref{fig:attn_dependency_switch}, right), the distribution shifts to one of non-uniformity. We consider the addition of any such turn that causes a switch from uniform to non-uniform or vice-versa in the creation of our visuals.

More formally, there is an attention dependency variable change from turn $T_i$ to $T_{i+1}$ with some threshold $\tau_a$ (note that $\tau_a=\tau$ in section~\ref{subsec:measuring_uniformity}) if any one of the following occurs:

\begin{enumerate}
\item $\alpha(\mathbf{w}_{i+1}) \ge \tau_a$ and $\alpha(\mathbf{w}_{i}) < \tau_a$
\item $\alpha(\mathbf{w}_{i}) \ge \tau_a$ and $\alpha(\mathbf{w}_{i+1}) < \tau_a$
\end{enumerate}

With \textbf{1}, we are switching from a uniform distribution to a non-uniform distribution with the addition of turn $T_{i+1}$. . With \textbf{2}, we are switching from a non-uniform distribution to a uniform distribution.

Note that it is possible that the attention dependency variable change is observed for many turns and not just one.

A \emph{context dependency switch} occurs when the addition of a turn causes a previous turn's weight to change significantly. In Figure~\ref{fig:context_dependency_switch}, the addition of turn $6$ causes turn $3$'s weight to spike.

Mathematically, there is a context dependency variable change in turn $T_j$ by addition of turn $T_{i+1}$ for $j < i+1$ with some threshold $\tau_c > 0$ if \[\left| \mathbf{w}_{i+1}[j] - \mathbf{w}_{i}[j] \right| \ge \tau_c.\] 

The final switch of consideration is a \emph{variation dependency switch}, which occurs when the weight of turn $i$ changes significantly over the \textit{entire} course of a conversation. 

More formally, there is a variation dependency variable change in turn $T_i$ with some threshold $\tau_v > 0$ when the conversation has $N$ turns if \[\frac{1}{N-i}\sum_{k=i}^{N-1} \big| \mathbf{w}_k[i] - \mathbf{w}_{k+1}[i] \big| \geq \tau_v\].

Note that variation dependency differs from context dependency as the latter determines turn $i$'s change with the addition of only one turn.

\begin{figure}[t]
\centering \includegraphics[scale = .35]{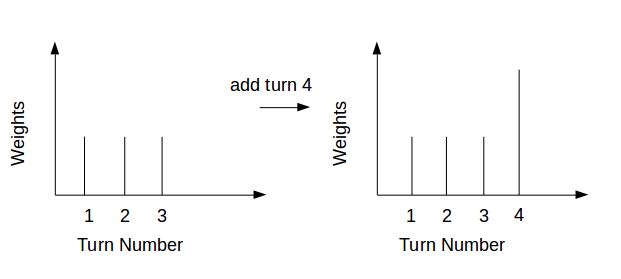}
\caption{Adding turn $4$ caused an attention dependency switch change because the distribution of weights switched from uniform to non-uniform. However, no turn previous to the added turn (turn $4$) changed significantly (no context dependency switch change).}
\label{fig:attn_no_context}
\end{figure}

\begin{figure}[t]
\centering \includegraphics[scale = .3]{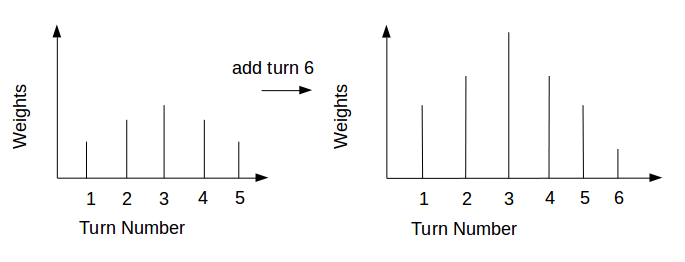}
\caption{Adding turn $6$ did not change the distribution of weights very much (no attention dependency switch). However, all weights previous to turn $6$ changed significantly (context dependency change).}
\label{fig:context_no_attention}
\end{figure}

For determining attention dependency, we considered normalized attention weights, but for variation and context, we considered the unnormalized output logits from the \han. It is also important to note that an attention dependency switch can occur without a context dependency switch and vice-versa. In Figure~\ref{fig:context_dependency_switch}, neither distribution is uniform; therefore, no attention dependency switch occurred. In Figure~\ref{fig:attn_no_context}, an attention dependency switch has occurred (uniform to non-uniform distribution), but there is no context dependency variable change. In Figure~\ref{fig:context_no_attention}, a context dependency variable change has occurred as many previous weights have spiked, but the distribution of weights has not changed (no attention dependency variable change bc it is still non-uniform).

In our experiments, we compute the thresholds mentioned in the definitions above as follows:
\begin{enumerate}

\item  For attention dependency, we experimented with various $\tau_a$ thresholds and tagged $100$ randomly chosen conversations for each of those thresholds to determine a potential candidate. For example, using a threshold of $.5$, weight vectors such as $[.2,.2,.6]$ would be considered uniform, which we greatly disagreed with. However, we determined that weight distributions below the $0.18$ threshold appeared uniform $90$\% of the time, which we considered good agreement.

\item For context dependency and variation dependency switches, we chose the value of $\tau_c=0.095$ and $\tau_v=0.124$, respectively, using the $75$th percentile of the values for different turns. Upon comparison with manual tagging of $100$ randomly chosen conversations, we agreed on all $100$ cases for the context dependency switch and $99$ out of $100$ cases for the variation dependency switch.  
\end{enumerate}

\subsection{Data and Classifier}

Our escalation data was obtained from~\cite{freemanbeaver2017}, which consists of  $7,754$ conversations ($20,808$ user turns) from two commercial airline IVAs. $1,268$ of the $7,754$ conversations had been tagged for escalation. See 
dataset statistics in Table~\ref{tbl:turn_stats_table}.

\begin{table}[H]
\centering
{
\begin{tabular}{|c|c|c|c|c|}
    \hline
    & \textbf{All} & \textbf{Airline $1$} &  \textbf{Airline $2$}\\ \hline
    \textbf{Min} & $1$ & $3$ & $1$\\ \hline
    \textbf{Q1} & $1$ & $3$ & $1$\\ \hline
    \textbf{Median} & $3$ & $4$ & $1$\\ \hline
    \textbf{Mean} & $3.38$ & $4.67$ & $1.43$\\ \hline
    \textbf{Q3} & $4$ & $5$ & $2$\\ \hline
    \textbf{95th per.} & $8$ & $9$ & $3$\\ \hline
    \textbf{Max} & $43$ & $43$ & $19$\\ \hline
\end{tabular}
}
\caption{Statistics on the number of user turns per conversation.}
\label{tbl:turn_stats_table}
\end{table}

Airline dataset $1$ has $2,998$ conversations and $14,000$ turns, and airline dataset $2$ has $4,756$ conversations and $6,808$ turns. The low turn counts present in dataset $2$ are due to the FAQ focus of dataset $2$'s particular IVA. Users tend to perform single queries such as ``baggage policy" instead of engaging in a conversational interaction.  In contrast, dataset $1$ originated from a more ``natural" IVA, and, therefore, users appeared to engage with it more through conversation.

The classifier (\han) used for escalation prediction is outlined in~\cite{yang2016hierarchical}. As the code was unavailable, we implemented \han with TensorFlow \cite{tensorflow2015-whitepaper}. Our version has substantially the same architecture as in \cite{yang2016hierarchical} with the exception that LSTM cells are used in place of GRU. We used the $200$-dimensional word embeddings from \textit{glove.twitter.27B} \cite{pennington2014glove} and did not adapt them during the training of our model. Each recurrent encoding layer has $50$ forward and $50$ backward cells, giving $100$-dimensional embeddings each for turns and conversations.

In predicting escalation, our network obtained an $F_1$ of $81.31\pm0.94 \%$ ($87.71\pm3.17\%$ precision, $75.90\pm2.61\%$ recall, averaged over five random splits). To compute these metrics, turn-level annotations were converted to conversation-level annotations by labeling a conversation as escalate if any turn in the conversation was labeled escalate.

For the visualization experiments, a random $80$-$20$ split was used to create training and testing sets. The training set consisted of $6,203$ conversations of which $1,027$ should escalate. The testing set consisted of $1,551$ conversations of which $241$ should escalate.

\subsection{Creating Our Visuals}
\label{subsec:create_visuals}
Given the occurrences of attention ($\mu$), context ($\beta$), and variation ($\gamma$) dependency switches, we now discuss how a visual of the entire conversation can be created. For each turn $T_i$, create a vector $v_i = [\mu_i, \beta_i, \gamma_i]$, where each variable inside this vector takes the value $1$ when the attention, context, and variation dependency switches trigger, respectively, and $0$ otherwise. 

Compute $\bar{v_i} = (\mu_i + \beta_i + \gamma_i)/3$, and use this value to represent the intensity of a single color (blue in our examples). The higher the value of $ \bar{v_i}$, the higher the color intensity.  Note that $\bar{v_i} \in \{0,\frac{1}{3},\frac{2}{3},1\}$. Take, for example, Table~\ref{tbl:pitbull_example} where for the first conversation's weights (using our weights), turns 2,3, and 6 have values of $\frac{2}{3}$, turns 4,5, and 7 have values of $\frac{1}{3}$, and the first turn has a value of $0$. Considering a higher dimension for $v_i$ which would create more values for $ \bar{v_i}$ is an objective for future work. 

\begin{table}[h]
\centering
\small{

{
\begin{tabular}{|p{0.6cm}|p{3.5cm}|p{0.9cm}|p{0.9cm}|}
    \hline
    \textbf{Turn} & \textbf{User Text} &  \textbf{\han Weight} & \textbf{Our Weight} \\
    \hline
    \textbf{1} & petsafe & \cellcolor{blue0}0.33 & 0.0\\ \hline

    \textbf{2} & contact live animal desk & \cellcolor{blue0}0.33 & \cellcolor{blue1}0.66\\ \hline

    \textbf{3} & petsafe & \cellcolor{blue0}0.33 & \cellcolor{blue1}0.66\\ \hline

    \textbf{4} & pit bull kennels \# ggkk98 & \cellcolor{blue0}0.33 & \cellcolor{blue0}0.33\\ \hline

    \textbf{5} & kennel requirements & \cellcolor{blue0}0.33 & \cellcolor{blue0}0.33\\ \hline

    \textbf{6} & live animal embargos & \cellcolor{blue0}0.33 & \cellcolor{blue1}0.66\\ \hline

    \textbf{7} & purchasing an in-cabin kennel & \cellcolor{blue0}0.33 & \cellcolor{blue0}0.33\\ \hline

\end{tabular}
}

\vspace{.5cm}

{
\begin{tabular}{|p{0.6cm}|p{3.5cm}|p{0.9cm}|p{0.9cm}|}
    \hline
    \textbf{Turn} & \textbf{User Text} &  \textbf{\han Weight} & \textbf{Our Weight} \\
    \hline

    \textbf{1} & is there a customer service phone number & \cellcolor{blue0}0.33 & \cellcolor{blue0}0.33\\ \hline

    \textbf{2} & i just requested a refund thru expedia cause i picked the wrong flight day , how long will it take ? & \cellcolor{blue0}0.33 & \cellcolor{blue0}0.33\\ \hline

    \textbf{3} & is there a way to expidite that ? & \cellcolor{blue0}0.33 & \cellcolor{blue1}0.66\\ \hline

    \textbf{4} & can we rush the refund cause i need to book another ticket & \cellcolor{blue0}0.33 & \cellcolor{blue1}0.66\\ \hline

    \textbf{5} & refunds & \cellcolor{blue0}0.33 & \cellcolor{blue2}1.0\\ \hline

    \textbf{6} & check refund status
 & \cellcolor{blue0}0.33 & \cellcolor{blue2}1.0\\ \hline

    \textbf{7} & refund processing times & \cellcolor{blue0}0.33 & \cellcolor{blue0}.33\\ \hline

\end{tabular}
}

\caption{The influence on escalation of each user turn in two sample conversations where 4 weight values are possible ($\{0,\frac{1}{3},\frac{2}{3},1\}$).}
\label{tbl:pitbull_example}
}
\end{table}

\section{Results and Discussion}
\label{sec:results_and_discussion}

We first considered the frequency of each of the behaviors discussed in section~\ref{subsec:attn_behaviors} as well as their co-occurrences with escalation.

After removing single turn conversations (as they are uniform by default), the number of turns that had a context dependency switch as a result of adding a new turn was $4,563$. However, the number of times that such an event coincided at least once with escalation was $766$. As it appeared that the effect of context dependency was quite low, we next considered the variation and attention dependency variables. The total number of turns that had a variation dependency switch was $2,536$, and $1,098$ also coincided with a change of escalation, indicating that a variation dependency switch is potentially valuable in the creation of new visuals. In addition, the number of uniform to non-uniform turn pairs (uniform weight distribution for first $i$ turns but non-uniform for first $i+1$ turns) was $1,589$ whereas the number of non-uniform to uniform turn pairs was $259$. Out of the times when there was a uniform to non-uniform switch, $710$ cases coincided with escalation compared to only $22$ for non-uniform to uniform changes. 
 
\begin{table}[h]
\centering
\small{
{
\begin{tabular}{|p{0.6cm}|p{4.2cm}|p{1.0cm}|}
    \hline
    \textbf{Turn} & \textbf{User Text} &  \textbf{Our Weight} \\ \hline
    \textbf{1} & where do you check to see about a refund that AIRLINE is giving us for a portion of our travel we did not get ? & 0.0 \\ \hline
    \textbf{2} & check refund status & \cellcolor{blue_uniform} 0.33 \\ \hline
    \textbf{3} & refund processing times & \cellcolor{blue_uniform} 0.33 \\ \hline

    \textbf{4} & check refund status & \cellcolor{blue2} 1.0 \\ \hline
    \textbf{5} & check refund status & \cellcolor{blue2} 1.0 \\ \hline
\end{tabular}
}
\caption{Example of a visual generated by our method which the $3$ reviewers tagged highly (a $9$ and two $10$s).}
\label{tbl:sample_tag}
}
\end{table} 
 
As shown in Figure~\ref{fig:motivation}, the use of our method when the \han weights are uniform greatly reduces or even eliminates the uniformity at lower $\alpha$ thresholds.  To determine if our visuals were also assigning weights properly, we had three reviewers rate on a $0$ to $10$ scale ($0$ being poor, $10$ being best) of how well each visualization highlights the influential turns for escalation in the conversation. See Table~\ref{tbl:sample_tag} for an example that was tagged nearly perfectly by reviewers.  As our method only serves to highlight influential turns in situations when the existing attention weights are uniform, no direct comparison was done to \han weights over the entire dataset.

To avoid bias, the chosen reviewers had never used the specific IVA and were not familiar with its knowledge base although they may have performed similar tagging tasks in the past. The annotators were reminded that if a turn is given a darker color, then that turn supposedly has greater influence in determining escalation. They were, thus, given the task of determining if they agree with the visualization's decision. A rating of $0$ was instructed to be given on complete disagreement, and $10$ upon perfect agreement. Consider a human representative given Our Weight in Table~\ref{tbl:comparison}, which highlights turn $4$ as the most influential turn on escalation, as opposed to the \han Weight which requires careful reading to make this determination.

From the $1,268$ conversations that escalated in the dataset, we first filtered conversations by a uniformity threshold, $\alpha=0.18$ (user chosen as described in subsection~\ref{subsec:attn_behaviors}). At this threshold, $10.9\%$ or $138$ conversations remained. Next, we filtered the conversations that were not correctly classified by \han, leaving $85$ or $6.7\%$.  

The average $0-10$ rating between the three reviewers over the remaining conversations was $6$. This demonstrates that on average, reviewers felt that the visualizations were adequate.  Put in perspective, adding adequate visuals to the thousands of daily escalations that would otherwise have no visual is a great improvement. 

In cases of uniform attention weights at the stopping point, this can also make it difficult to spot potential areas for classifier improvement if we do not incorporate turn weight fluctuations as the conversation progresses to the stopping point. For example, in the first escalated conversation displayed in Table~\ref{tbl:pitbull_example}, turn 6 has a high weight under our scheme because of the presence of the word ``live". Customers will frequently ask for a  ``live customer representative" which is a sign for escalation. However, in Table~\ref{tbl:pitbull_example}, ``live" is used in a different context, but the weight given to it is high due to turn weight fluctuations as the conversation progresses to the stopping point. Our weights expose this potential problem for the classifier which may suggest using n-grams or some other methodology for improvement. If we were to use uniform \han weights at the stopping point only, we might miss these areas for improvement.

In addition to the possible reduction in human review time and spotting potential areas for classifier improvement, the visuals only required $0.9$ milliseconds on average to compute per conversation (on a laptop with an Intel Core i7-4710MQ CPU @ 2.50GHz, 16 GB of RAM, running Ubuntu 16.04). This adds insignificant latency to the transfer while generating the visualization, which is an important goal.

In the future, this work would greatly benefit from an expanded dataset. As we only wish to consider conversations with uniform weights on the turn of escalation, this cuts our dataset dramatically, necessitating a larger tagged dataset. Considering more attention behaviors so we can have higher granularity of color intensity is also an objective of future work. As our method only looks at the changes in attention weight, our method is not task-specific.  Therefore, it would be beneficial to test our methodology on visualizing other sequential analysis tasks besides escalation, such as fraud or anomaly detection or applications in the medical domain~\cite{velupillai2015towards,martinez2015automatic}.

\section{Conclusion}

Although attention in deep neural networks was not initially introduced to inform observers, but to help a model make predictions, it can also be used to inform.  In the instances where a model thinks all historical samples should be considered equally important in a sequential analysis task, we must look elsewhere for a computationally inexpensive means to understand what happened at the stopping point.  In this paper, we have introduced such a means by monitoring attention changes over the sequential analysis to inform observers.  This method introduces negligible overhead, an important consideration in real-time systems, and is not tied to the implementation details or task of the model, other than the prerequisite of an attention layer.

\bibliography{freemanblackbox2018}
\bibliographystyle{acl_natbib_nourl}

\end{document}